\documentclass[10pt,twocolumn,letterpaper]{article}

\usepackage{iccv}
\usepackage{times}
\usepackage{epsfig}
\usepackage{graphicx}
\usepackage{amsmath}
\usepackage{amssymb}
\usepackage{enumitem}
\usepackage{booktabs}
\usepackage{colortbl}
\usepackage[most]{tcolorbox}
\usepackage{pifont}
\usepackage{tikz,pgfplots}
\usepackage{multirow}

\usepackage{hyperref}
\hypersetup{breaklinks=true,letterpaper=true,colorlinks,}

\newcommand{\myparagraph}[1]{\noindent \textbf{#1} --}
\newcommand{\code}[1]{{\small\texttt{#1}}}
\newcommand{\scriptcode}[1]{{\scriptsize\texttt{#1}}}

\definecolor{cvprblue}{rgb}{0.21,0.49,0.74}
\definecolor{TableGray1}{HTML}{9B9B9B}
\definecolor{TableGray2}{HTML}{C0C0C0}
\definecolor{TableGray3}{HTML}{EEEEEE}
\definecolor{colorig}{HTML}{E8EEE4}
\definecolor{coloriig}{HTML}{FAF0E9}
\definecolor{colorog}{HTML}{FFFBEC}
\definecolor{colorlegendcol}{HTML}{e9f8fb}
\definecolor{colordebit}{HTML}{DBF9DB}
\definecolor{colortransfer}{HTML}{FAF0E9}
\definecolor{colorprobing}{HTML}{FFFBEC}
\newcommand{\yes}{\textcolor{black}{\ding{51}}}
\newcommand{\no}{\textcolor{TableGray1}{\ding{55}}}

\newcommand{\wscratch}{$\circlearrowright$}
\newcommand{\wfalse}{$f$}
\newcommand{\wtrue}{$t$}
\newcommand{\wft}{$\rightarrow$}
\newcommand{\wfr}{$\ast$}
\newcommand{\wpt}{$p$}

\definecolor{colorslideyes}{HTML}{FAF0E9}
\definecolor{colorslideyesleg}{HTML}{F7B17C}
\definecolor{colorslideno}{HTML}{E8EEE4}
\definecolor{colorslidenoleg}{HTML}{BFEE9A}
\definecolor{colorpretrain}{HTML}{e9f8fb}
\definecolor{colorpretrainleg}{HTML}{AFECFC}
\definecolor{colorfinetune}{HTML}{F1EFE6}
\definecolor{colorfinetuneleg}{HTML}{DBC78B}
\newcommand{\supmat}{Supplementary\,}
\newcommand{\supmatp}{Supplementary}

\definecolor{colfinetune}{HTML}{faf9d4}
\definecolor{coltrain}{HTML}{fad4e8}
\definecolor{colfreeze}{HTML}{d4d5fa}

\newcommand{\boxfinetune}[1]{\tcbox[on line,colframe=white,boxsep=0pt,left=1pt,right=1pt,top=0pt,bottom=0pt,colback=colfinetune]{#1}}
\newcommand{\boxtrain}[1]{\tcbox[on line,colframe=white,boxsep=0pt,left=1pt,right=1pt,top=0pt,bottom=0pt,colback=coltrain]{#1}}
\newcommand{\boxfreeze}[1]{\tcbox[on line,colframe=white,boxsep=0pt,left=1pt,right=1pt,top=0pt,bottom=0pt,colback=colfreeze]{#1}}

\newcommand{\boxslideyes}[1]{\tcbox[on line,colframe=white,boxsep=0pt,left=1pt,right=1pt,top=0pt,bottom=0pt,colback=colorslideyesleg]{#1}}
\newcommand{\boxslideno}[1]{\tcbox[on line,colframe=white,boxsep=0pt,left=1pt,right=1pt,top=0pt,bottom=0pt,colback=colorslidenoleg]{#1}}
\newcommand{\boxpretrain}[1]{\tcbox[on line,colframe=white,boxsep=0pt,left=1pt,right=1pt,top=0pt,bottom=0pt,colback=colorpretrainleg]{#1}}
\newcommand{\boxtransfer}[1]{\tcbox[on line,colframe=white,boxsep=0pt,left=1pt,right=1pt,top=0pt,bottom=0pt,colback=colorfinetuneleg]{#1}}

\DeclareRobustCommand\sampleline[1]{%
  \tikz\draw[#1] (0,0) (0,\the\dimexpr\fontdimen22\textfont2\relax)
  -- (2em,\the\dimexpr\fontdimen22\textfont2\relax);%
}

\definecolor{ColComments}{HTML}{e1945b}
\newcommand{\rem}[1]{\texttt{\scriptsize \textcolor{ColComments}{// #1}}}

\begin{document}

\title{What does really matter in image goal navigation?}
\author{Gianluca Monaci, Philippe Weinzaepfel, Christian Wolf\\
NAVER LABS Europe, Grenoble, France\\
{\tt\small firstname.lastname@naverlabs.com}
}

\maketitle

\begin{abstract}
\noindent
Image goal navigation requires two different skills: firstly, core navigation skills, including the detection of free space and obstacles, and taking decisions based on an internal representation; and secondly, computing directional information by comparing visual observations to the goal image. Current state-of-the-art methods either rely on dedicated image-matching, or pre-training of computer vision modules on relative pose estimation. In this paper, we study whether this task can be efficiently solved with end-to-end training of full agents with RL, as has been claimed by recent work. A positive answer would have impact beyond Embodied AI and allow training of relative pose estimation from reward for navigation alone. In this large experimental study we investigate the effect of architectural choices like late fusion, channel stacking, space-to-depth projections and cross-attention, and their role in the emergence of relative pose estimators from navigation training. We show that the success of recent methods is influenced up to a certain extent by simulator settings, leading to shortcuts in simulation. However, we also show that these capabilities can be transferred to more realistic setting, up to some extent. We also find evidence for correlations between navigation performance and probed (emerging) relative pose estimation performance, an important sub skill. 
\end{abstract}

\section{Introduction}
\label{sec:intro}
\noindent
The interplay between perception and action in modern AI-based robotics is a well-studied topic covering contributions from multiple fields, including computer vision, sequential decision making, and optionally some form of mapping and planning. Independent of the specific robotics task at hand, there is an ongoing debate on whether these components should be trained end-to-end, \eg with RL~\cite{CrocoNav2024,yadav2023ovrlv2}, typically for higher performance, or whether modular techniques could provide higher robustness, \eg~\cite{goatrss2024,krantz2023navigating}.

In the context of navigation, the \textit{ImageNav} task adds an additional complexity, as perception not only involves the currently observed image, but also the goal image, requiring the comparison of two images to extract \textit{directional} information. 
Different strategies have been proposed to address this, and the recent literature seems to provide conflicting reports on their performances and up- and downsides. 
We aim to shed some light on this with an in-depth analysis.

\begin{figure}[t] \centering
    \includegraphics[width=\linewidth]{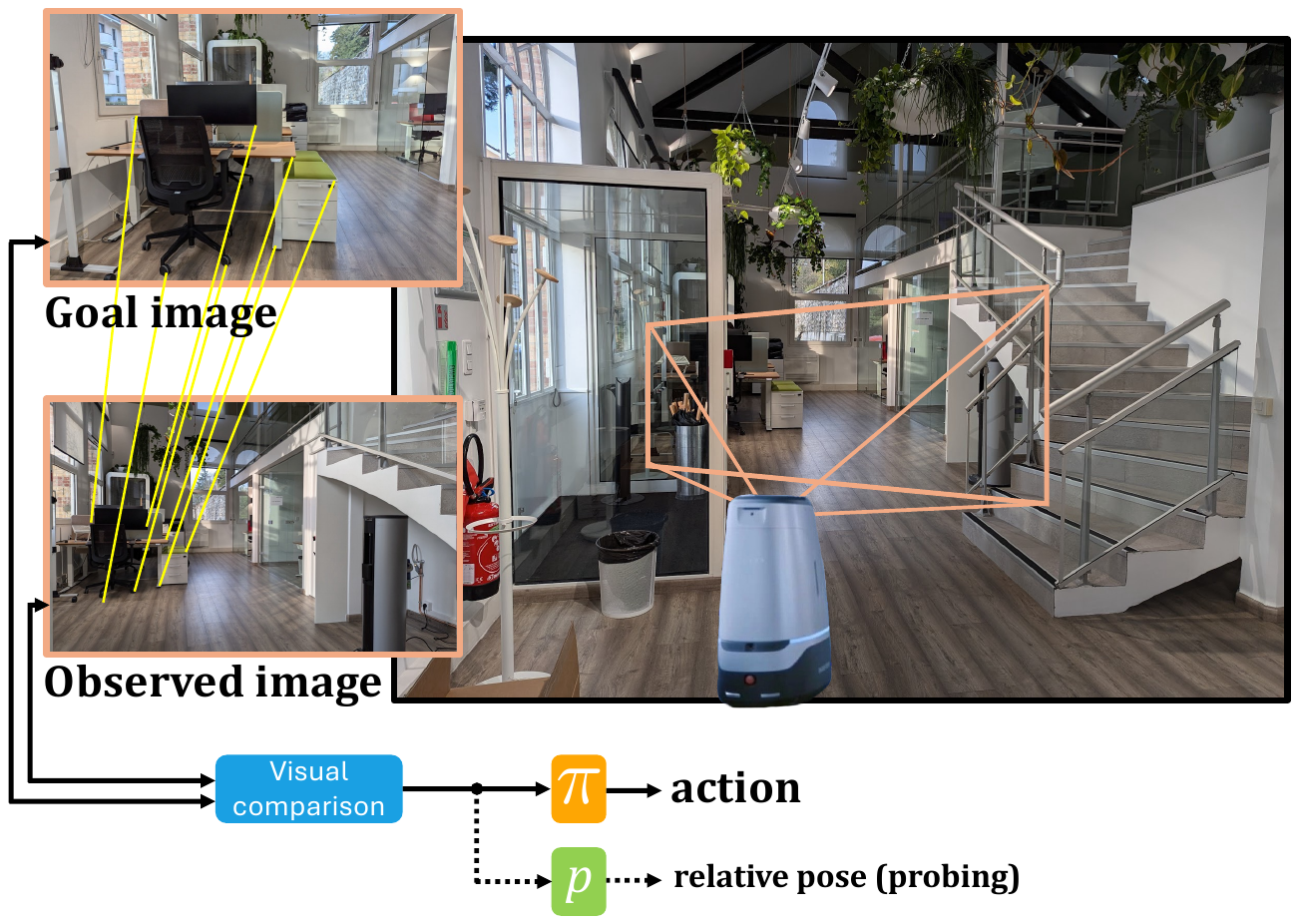}
    \caption{\label{fig:teaser}\textbf{Image goal navigation} requires general navigation skills, but also in particular the extraction of directional information towards the goal. We analyze which architecture design choices influence these capabilities, and to what degree they --- and the underlying sub task of relative pose estimation, which we probe with a dedicated head $p$ --- can be trained end-to-end from the navigation loss directly, without any pose ground-truth.}    
\end{figure}

Early works addressed the problem in a purely data-driven way by end-to-end training \textit{Late Fusion} (eg. \textit{Siamese}) image encoders \cite{zhu_target-driven_2017}, with mixed results. It was conjectured, that the weak and sparse learning signal from RL 
might not be sufficient to train high capacity encoders comparing images. Since the \textit{ImageNav} task introduces a sub task of relative pose estimation (\textit{``where is the goal with respect to my current position?''}), there is a strong case for modular methods separating perception and decision taking. For instance, Krantz~\etal \cite{krantz2023navigating} employ local features and matching (\textit{SuperGlue} \cite{SuperGlue2020}) combined with map and plan strategies for navigation. However, it has since been shown \cite{CrocoNav2024} that image correspondences can actually emerge directly in the cross-attention layers of a binocular transformer, provided it is appropriately pre-trained. This strategy, used in the \textit{DEBiT} agent
\cite{CrocoNav2024} outperforms explicit matching approaches.
Key to this solution is, firstly, pre-training, and secondly, the fusion of observation and goal
such that comparisons between local parts (\eg patches) of the two respective images can be calculated by the architecture, as opposed to comparing embedding vectors.

Recent work~\cite{sun2024fgprompt} however indicates that \textit{ImageNav} can be addressed by very low-capacity convolutional encoders with channel-wise stacking and through RL training alone, without any pre-training. \textit{FGPrompt}~\cite{sun2024fgprompt} achieves results which are on par with \textit{DEBiT} with a fraction of the model capacity, which raises interesting questions: 
\begin{description}[nosep,,itemsep=1mm,labelindent=0mm,leftmargin=3mm,topsep=1mm]
    \item[Q1:] Can \textit{ImageNav} be learned by RL alone?
    \item[Q2:] Which architecture can support that best?
    \item[Q3:] Is the solution related to the estimation of directional information between observation and goal images?
\end{description}
The implications could potentially go beyond image goal navigation and have impact on the wider field of relative pose estimation and localization. In this paper we systematically disentangle the different mediators to shed a light on what choices matter in \textit{ImageNav}, and we find that:
\begin{description}[nosep,,itemsep=1mm,labelindent=0mm,leftmargin=3mm,topsep=1mm]
\item[R1:] Agents trained with \textbf{RL alone underperform} in realistic navigation settings.
\item[R2:] \textit{Early Fusion} performs better, with \textbf{early patch-wise fusion being essential}, compared to \emph{Late Fusion}. 
\item[R3:] We find \textbf{correlations between navigation and (emergent) relative pose estimation performance}, which we probe from representations trained with RL. 
\item[Bonus:] We discovered that success of recent frugal architectures using channel stacking and trained with RL alone is \textbf{mostly due to a simulator setting that allows agents to slide along walls}. Surprisingly, while this setting is known to hamper sim2real transfer since \cite{kadian20sm2real}, we have obtained further insights: (i) unrealistic motion simulation has also a negative impact on the learned perception capabilities, and (ii) some learned capabilities can still be transferred to realistic settings \textbf{if the transfer includes weights of the perception module}.
\end{description}

\section{Related Work}
\myparagraph{Navigation from visual observations} has been addressed in several different fields. Robotics for a long time focused on explicit modeling~\cite{burgard1998interactive,macenski2020marathon,marder2010office}, which is mainly based on mapping and localization~\cite{bresson2017simultaneous, labbe19rtabmap,thrun2005probabilistic}, explicit planning~\cite{konolige2000gradient, sethian1996fast} and low-level control \cite{fox1997dynamic,rosmann2015timed}. Accurate sensor and observation models are essential, followed by filtering, dynamical models and optimization techniques. 

Modern AI-based solutions are typically trained on large-scale photorealistic simulators like Habitat \cite{Savva_2019_ICCV} or AI2-Thor~\cite{ai2thor}. Modular agents \cite{Chaplot2020Learning,ramakrishnan_poni_2022,raychaudhuri2024mopa} decompose the problem in sub modules, typically mapping, pose estimation, planning and local decision taking.  
On the other end of this spectrum, end-to-end trained models directly map input to actions, with RL~\cite{DBLP:conf/iclr/JaderbergMCSLSK17,mirowski17learning,zeng2024poliformer,SPIN2024} or Imitation Learning (IL)~\cite{DBLP:conf/nips/DingFAP19}, offline-RL \cite{perceiverAC2024}, or unsupervised RL \cite{kim2024unsupervisedRL,liu2025a}.
In recent work, end-to-end trained agents are combined with visual foundation models, eg.  \textit{DINOv2} \cite{oquab2023dinov2} in \textit{PoliFormer} \cite{zeng2024poliformer}, or binocular ones for image comparisons as in DEBiT \cite{CrocoNav2024}.
Agent memory is implemented as recurrent representations~\cite{FMNav09CVPR,janny2025reasoning}, occupancy maps~\cite{Chaplot2020Learning}, semantic maps~\cite{chaplot2020object}, latent metric maps~\cite{DBLP:conf/pkdd/BeechingD0020,Henriques_2018_CVPR,DBLP:conf/iclr/ParisottoS18}, topological maps~\cite{BeechingECCV2020,Chaplot_2020_CVPR,shah2023vint}, scene graphs \cite{singh2022generalpurposesupervisorysignal}, explicit episodic memory~\cite{chen_think_2022,DuICLR21VTNetVisualTransformerNetwork4ObjectGoalNavigation,Fang_2019_CVPR,reed_generalist_2022}, implicit representations~\cite{Marza2022NERF,kwon2023renderable} or navigability~\cite{Mole2024}.
In this work we investigate end-to-end trained agents with recurrent memory.

\myparagraph{Image goal navigation}
or ``\textit{ImageNav}'', adds a matching aspect to navigation, as the agent needs to compare the goal image to the observed image. 
Explicit methods have addressed this with local feature matching \cite{krantz2023navigating}, or by retrieving 
features from a topological map \cite{BeechingECCV2020}. End-to-end trained agents compare images by extracting features with ResNets \cite{al2022zero,zhu_target-driven_2017} or ViTs \cite{yadav2023ovrlv2}, potentially stacking multiple observations over time \cite{shah2023vint}, with ViTs followed by cross-attention \cite{CrocoNav2024}, or by stacking images channel-wise \cite{sun2024fgprompt}.
Modular approaches have also been proposed~\cite{DasNeuralModularControl2018,wu22image_goal}. 

Depending on whether goal images are randomly chosen or correspond to images of semantically meaningful objects, the task can either be supported by semantic features or requires purely geometric image understanding. In this work we focus on the latter, and study how success depends on architectural key design choices. We review these choices in detail in \cref{sec:archchoices} and their implementations in the state of the art in \cref{sec:implementationsinsota}.

\myparagraph{Relative pose estimation (RPE)} was tackled for decades with pixel-level image matching techniques~\cite{mvgeo,orbslam,sfmrevisited}. Learning-based approaches have also been proposed~\cite{posenet}, and to overcome the lack of generalizability of early methods, self-supervised techniques have been used~\cite{undeepvo}. More recently, \emph{DUSt3R}~\cite{wang2024dust3r} regresses pointmaps of each image expressed in the coordinate system of the first image while \emph{MASt3R}~\cite{mast3r} additionally learns an extra descriptor inspired by standard image matching. Both leverage \emph{CroCo} pre-training~\cite{CroCo2022,CroCoV2}. Similarly, \emph{MicKey}~\cite{mickey} regresses pointmaps of each image in its respective coordinate system and learns descriptors to obtain correspondences, supervised by relative pose ~\cite{reinforcedpoints,dsac}.
These recent methods have led to impressive results, even under scenarios with little image overlap as in the MapFree-Relocalization benchmark~\cite{mapfree}. In this work, we study RPE as a sub-task of ImageNav and probe it explicitly from agent representations.

\begin{figure*}[t] \centering
\includegraphics[width=0.8\linewidth]{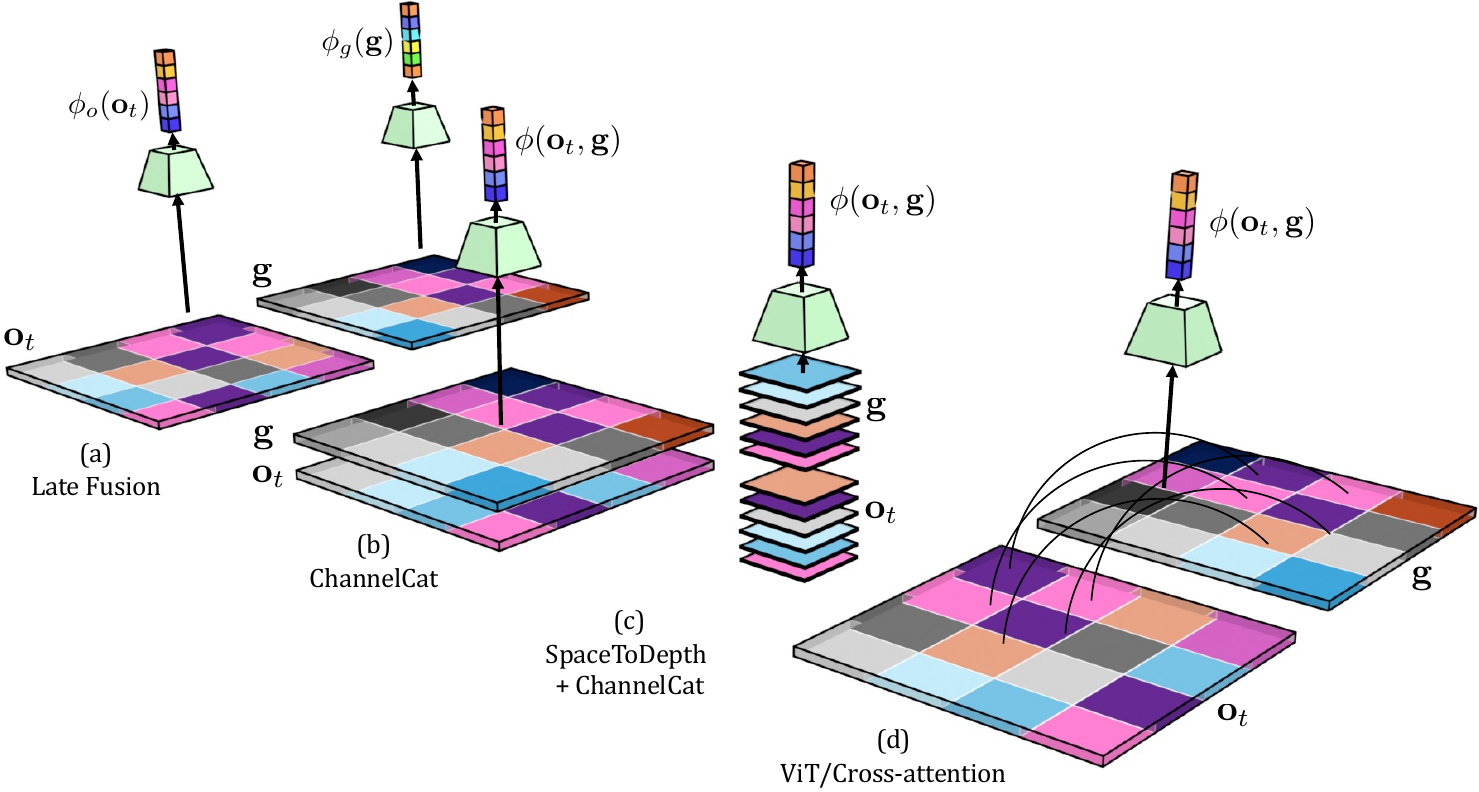} \\[-0.1cm]
\caption{\label{fig:arch} \textbf{Different architecture choices for binocular encoders} learning to compare the observed image $\mathbf{o}_t$ with the goal image $\mathbf{g}$: (a) \textit{Late Fusion} encodes them separately and comparison is done ``\textit{late}'' between embedding vectors $\phi_o(\mathbf{o}_t)$ and $ \phi_g(\mathbf{g})$, making correspondence computations difficult. (b) \textit{ChannelCat} stacks images over the channel dimension, followed by convolutional encoders $\phi\left([\mathbf{o}_t, \mathbf{g}]_{\textrm{dim}=1}\right)$. It makes correspondence computations possible in principle if the CNN receptive field is big enough. (c) \textit{SpaceToDepth} reshapes the patch dimension into the channel dimension. Combined with \textit{ChannelCat} \cite{sun2024fgprompt}, it could allow correspondence to emerge in each layer directly through conv filters. (d) \textit{Binocular ViTs} \cite{CrocoNav2024} model correspondence directly as cross-attention between patch tokens.
}
\end{figure*}

\section{Can we learn goal-perception with actions?}
\noindent
We study the \textit{ImageNav} task in photo-realistic 3D environments, where an agent is given a goal image $\mathbf{g}{\in} \mathbb{R}^{3{\times}H{\times}W}$ and is required to
navigate from a starting location to the position shown in the goal. At each timestep $t$ the agent observes an RGB image $\mathbf{o}_t{\in}\mathbb{R}^{3{\times}H{\times}W}$, all images are of size $112{\times}112$.
The action space is the discrete set $\mathcal{A}$  =\{{\tt \small MOVE FORWARD 0.25m}, {\tt \small TURN LEFT $10^{\circ}$}, {\tt \small TURN RIGHT  $10^{\circ}$}, and {\tt \small STOP}\}. An episode is considered successful if the agent is within 1m of the goal position \textit{and} it calls the {\tt \small STOP} action within its 1000 steps budget. We use the Habitat simulator \cite{Savva_2019_ICCV} and the Gibson dataset \cite{xiazamirhe2018gibsonenv}.

All our experiments are done with different variants of the same identical base agent, which maintains a recurrent episodic memory $\mathbf{h}_{t}$, integrates the observation $\mathbf{o}_t$ and the goal image $\mathbf{g}$, and predicts actions $\mathbf{a}_t$:
\begin{equation}
    \arraycolsep=1.4pt
    \begin{array}{lll}
    \tilde{\mathbf{g}}_t & = \phi(\mathbf{o}_t, \mathbf{g}) &
    \rem{binocular perception}
    \\
    \mathbf{h}_t & = h(\mathbf{h}_{t-1}, \tilde{\mathbf{g}}_t, \zeta(\mathbf{a}_{t-1})) &
    \rem{state update}
    \\
    \mathbf{a}_t & \sim \pi(\mathbf{h}_t), &
    \rem{policy}\\
    \end{array}
    \label{eq:agent2}
\end{equation}
where $h$ is the function updating the hidden state $\mathbf{h}_t$ of a GRU \cite{cho-etal-2014-learning} and gating equations have been omitted for brevity; 
$\phi$ and $\zeta$ are trainable encoders, and $\pi$ is a linear policy. All agent variants are trained from scratch with PPO~\cite{schulman2017proximal} with reward function defined as in~\cite{chattopadhyay2021robustnav,CrocoNav2024}:
$
r_t=\mathrm{K} \cdot \mathbf{1}_{\text {success}}-\Delta_t^{\mathrm{Geo}}-\lambda ,
\label{eq:rewardn}
$
where $\lambda{=}0.01$ is a slack cost to encourage efficiency, $K{=}10$, and $\Delta_t^{\mathrm{Geo}}$ is the increase in geodesic distance to the goal.

\subsection{Architecture design choices}
\label{sec:archchoices}

\noindent
At the heart of the problem is the sub task of learning to compare images $\mathbf{o}_t$ and $\mathbf{g}$ to infer directional information related to relative pose estimation. In the classical \textit{ImageNav} task, goal images can be taken from any position and do not necessarily show a semantically meaningful object. Given the possible absence of semantic cues, which would have allowed to compare images on a global level, we conjecture that successful comparison requires the computation of some form of correspondence between local parts of the respective images. 
While pose estimation methods based on local feature matching perform this explicitly \cite{R2D2_2019,SuperGlue2020}, 
recent work have shown 
that this kind of correspondence can emerge implicitly from large-scale training without explicit supervision of matches, \eg in \textit{CroCo} \cite{CroCo2022}
or \textit{DUSt3R} \cite{wang2024dust3r}. 

In the context of robotics and navigation, the question arises whether it is possible to go a step further, and forgo pose-related losses completely, training image comparisons end-to-end together with the navigation agent from RL alone. This has been the promise of works like 
\textit{FGPrompt} \cite{sun2024fgprompt} and \textit{OVRL-v2} \cite{yadav2023ovrlv2}, and we will analyze this in-depth.

Assuming that correspondence computations are necessary to extract directional information, we conjecture that these computations need to be supported by underlying network architectures. 
The basic building blocks for image comparison in the literature, also illustrated in \cref{fig:arch}, are:
\begin{description}[nosep,,itemsep=1mm,labelindent=0mm,leftmargin=3mm,topsep=1mm]
    \item[Late Fusion networks] use separate networks $\phi_o$ and $\phi_g$ to encode observation and goal. The representation fed to the agent is given as $\phi(\mathbf{o}_t,\mathbf{g})=[ \phi_o(\mathbf{o}_t),\phi_g(\mathbf{g})]$, where $[.]$ denotes concatenation, see \cref{fig:arch}(a).     
    The comparison is thus performed ``\textit{late}'' between embeddings $\phi_o(\mathbf{o}_t)$
    and $\phi_g(\mathbf{g})$, which makes it generally harder to be done on a local image level, unless output representations retain sufficient spatial structure going through encoder layers. 

    \item[ChannelCat] 
    uses a single network to encode both observation and goal, which are channel stacked into one input image
    $\phi\left([\mathbf{o}_t, \mathbf{g}]_{\textrm{dim}=1}\right)$, where 0 is the batch dimension and 1 the channel dimension --- see \cref{fig:arch}(b).

    If encoders are implemented as CNNs, a common choice in robotics for computational complexity reasons, comparisons between local parts of 
    images can be done only if the receptive field of the network is big enough to encompass the full image. Given a sufficiently deep encoder correspondence can therefore be computed, but not necessarily efficiently, as individual matching results need to be accumulated in the output of different filter kernels.

    If encoders are ViTs, each feature patch corresponds to a pair of patches having the same coordinates in both images. Correspondence can be more easily computed if different self-attention heads are learned with query and key projections focusing on one of the two channel groups of the two respective images.\footnote{As an illustrative example, consider an input image $X$ corresponding to channel stacking of two images $X_A$ and $X_B$. Then, if for a given attention head the Query projection only uses the channels of image $X_A$ and the Key projection only use the channels of image $X_B$, then this self-attention on $X$ mimics cross-attention between $X_A$ and $X_B$.}
    
    \item[SpaceToDepth] reshapes image patches into channel values 
    \cite{ridnik2021tresnet}. \textit{FGPrompt} \cite{sun2024fgprompt} uses \textit{SpaceToDepth} in combination with \textit{ChannelCat} of observation and goal, achieving strong navigation results. We investigate whether a ResNet in this configuration can compute correspondences across a large spatial dimension through a single convolutional layer, somewhat reminiscent of cross-attention, only with very few parameters --- see \cref{fig:arch}(c). 
    
    \item[Cross-attention] is a natural way to compute correspondences between local parts of images \cite{CrocoNav2024}, as each patch in one image can be naturally linked to one or more patches in the other image through the cross-attention distribution, see \cref{fig:arch}(d).
\end{description}

\subsection{Implementations in prior work}
\label{sec:implementationsinsota}

\noindent
Early methods used \textit{Late Fusion} approaches based on CNNs \cite{al2022zero,zhu_target-driven_2017}, but later works  switched to ViTs.
Cross-attention for \textit{ImageNav} was introduced in \textit{DEBiT} \cite{CrocoNav2024} with code available,\footnote{\url{https://github.com/naver/debit}} and combined with pre-training, first for cross-view completion as in \cite{CroCo2022}, then for relative pose and visibility estimation. The binocular ViT was combined with an additional CNN with only the observation as input, trained with RL, providing current state-of-the-art results. Interestingly, the authors have shown that pre-training is essential, as the high-capacity binocular transformer cannot be trained from scratch with RL only.
The learning signal from the RL loss alone seems to be too weak 
to drive the encoder to discover correspondence computations. The inability of RL to train big encoders was somewhat confirmed in the \textit{Late Fusion} method \textit{OVRL-v2} \cite{yadav2023ovrlv2}, as ``\textit{switching from the 50.9M parameters ViT-Small to the 179.2M parameters ViT-Base produces another negative result: SR drops -2.4\%
while SPL minimally increases by +0.7\%}''.

\textit{FGPrompt}~\cite{sun2024fgprompt} proposed channel stacking for \textit{ImageNav}, with code available,\footnote{\url{https://github.com/XinyuSun/FGPrompt}} and it is based on previous code of \textit{ZSEL}~\cite{al2022zero}. Both codebases use a non-standard ResNet9 architecture with several custom blocks, the most significant one being \textit{SpaceToDepth} described earlier. Interestingly, \textit{SpaceToDepth} was first introduced in \textit{TResNet}~\cite{ridnik2021tresnet} as a general purpose network block for computer vision optimized for memory efficiency, targeting image classification and object detection. Correspondence problems were not explicitly targeted. The ResNet9 encoder in this paper is exactly the same as those used in \textit{ZSEL} and \textit{FGPrompt}.

Given the good results \textit{FGPrompt} obtained without any pre-training, 92\% SR, vs. \textit{DEBiT}'s 94\% SR relying on pre-training, we raise the question whether it is possible to use the weak learning signal provided by RL to learn binocular perception modules capable of providing directional information. A positive answer would have significant impact also on training vision modules targeting RPE directly. In the experimental section we will dive into this question and benchmark  architectures and design choices. 

\section{Experimental Setup}
\noindent
Unfortunately there are different evaluation protocols in the literature on Embodied AI, and in particular on the \textit{ImageNav} task. We therefore train all agents ourselves in identical conditions and with the same experimental protocol.
The agents are trained on the 72 scenes of the Gibson dataset \cite{xiazamirhe2018gibsonenv} and use the standard Habitat episode definitions (a subset of the literature uses the definitions from \textit{Episodic Transformers} \cite{MezghaniMemAug2021} and another subset those from \textit{NSNRL} \cite{NoSimNoRL2021}). We set the maximum episode length to the default value of 1000 steps and we train for 500M steps, which allows all agent variants to converge easily.

\myparagraph{Implementation details} We test perception networks $\phi$ implemented as \textit{ResNet9}, \textit{ViT-Small}\footnote{\url{https://github.com/huggingface/pytorch-image-models}} and \textit{DEBiT-B} architectures, available in their respective public repositories. We test \textit{CannelCat}, \textit{Late Fusion} and \textit{SpaceToDepth} variants of ResNet and ViT, while for \textit{DEBiT} we either use the pretrained binocular encoder or train the whole agent from scratch. Network versions are chosen based on size-performance tradeoff --- see \supmat for details.
The function $\zeta$ embeds previous actions into a 32D feature, $h$ is a GRU with 2 layers and hidden dimension 128, followed by a linear Actor-Critic policy $\pi$.

\myparagraph{Validation}
To avoid overfitting evaluation over the choice of checkpoints, while at the same time staying comparable, we use three different splits. We train on the standard Gibson train split (72 scenes) and report on the standard Gibson validation split (14 scenes), as most literature, but we use an independent custom holdout set of unused Gibson scenes to choose the checkpoint.

\myparagraph{Metrics}  Navigation performance is evaluated by success rate (SR), \ie, fraction of episodes terminated within a distance of ${<}1$m to the goal by the agent calling the \code{STOP} action, and SPL~\cite{AndersonX18OnEvaluationOfEmbodiedNavAgents}, \ie, SR weighted by the optimality of the path,
$
\textit{SPL}=\frac{1}{N} \sum_{i=1}^N S_i \frac{\ell_i^*}{\max (\ell_i, \ell_i^*)} ,
\label{eq:spl}
$
where $S_i$ be a binary success indicator in episode $i$, $\ell_i$ is the agent path length and $\ell_i^*$ the shortest path length.

\myparagraph{Comparing the comparable}
An important choice in the Habitat simulator is the binary \code{Sliding} switch, which, when set to \code{True}, allows the agent in simulation to slide along obstacles when colliding, against the more realistic behavior of stopping. 
This parameter is known to have a big impact on sim2real transfer~\cite{kadian20sm2real}. In our own analysis, we saw that the increased difficulty of \code{Sliding=False} goes beyond sim2real transfer and also impacts task difficulty in simulation: when the agent hits a wall, it can in principle continue the episode, but is frequently restricted in its motion, often stuck, and typically needs to turn to recover.

While this setting is switched to \code{True} by default, there is consensus in the field that it should be set to \code{False}, as the goal is to decrease the sim2real gap and to evaluate agents in situations which are as close as possible to real physical robots and environments. Looking closer into the available code of navigation repositories, it seems that indeed most use \code{Sliding=False} (\textit{DEBiT}~\cite{CrocoNav2024}, \textit{ZSEL}~\cite{al2022zero}, \textit{ZSON}~\cite{majumdar2022zson}, \textit{PIRLNav}~\cite{ramrakhya2023pirlnav}), with a couple of exceptions using \code{Sliding=True}, notably \textit{FGPrompt}~\cite{sun2024fgprompt} (details in \supmatp).
We therefore performed experiments with both settings, and were surprised to see that there is a big influence of this parameter. Given its impact, we color-coded the entries in all tables into three different groups: (i) \boxslideyes{\code{Sliding=True}}: the visual encoder $\phi$ has been trained with this setting on; (ii) \boxslideno{\code{Sliding=False}}: it has been trained with this setting off; (iii) \boxpretrain{Pre-train}: the visual encoder $\phi$ has been pre-trained on an auxiliary task, typically relative pose and visibility estimation (RPVE).

\begin{table}[t] \centering
{\small
    \setlength{\tabcolsep}{2pt}
    \begin{tabular}{clcl|rr|rr}
            \specialrule{1pt}{0pt}{0pt}
            \rowcolor{TableGray2} & &  & & \multicolumn{2}{c|}{\cellcolor{colorslideyesleg}\textbf{Slide=True}} & \multicolumn{2}{c}{\cellcolor{colorslidenoleg}\textbf{Slide=False}} \\ \cline{4-8}
\rowcolor{TableGray2} & \multirow{-2}{*}{\textbf{Model}} & \multirow{-2}{*}{\textbf{s2d$^\dagger$}} & \multirow{-2}{*}{\textbf{Backbone}} & \cellcolor{colorslideyesleg}\textbf{SR} & \cellcolor{colorslideyesleg}\textbf{SPL} & \cellcolor{colorslidenoleg}\textbf{SR} & \cellcolor{colorslidenoleg}\textbf{SPL} \\ 
            \specialrule{0.5pt}{0pt}{0pt}
            \rowcolor{TableGray3}
            (a)&
            Late Fusion &
            \no & 
            ResNet9 &
            \cellcolor{colorslideyesleg}13.8 & \cellcolor{colorslideyesleg}8.0 &
            \cellcolor{colorslidenoleg}12.8 & \cellcolor{colorslidenoleg}7.1
            \\
            \rowcolor{TableGray3}
            (b)&
            \cellcolor{TableGray3}Late Fusion \cite{al2022zero}&
            \cellcolor{TableGray3}\yes & 
            \cellcolor{TableGray3}ResNet9 &
            \cellcolor{colorslideyesleg}12.5 & \cellcolor{colorslideyesleg}7.6 &
            \cellcolor{colorslidenoleg}13.2 & \cellcolor{colorslidenoleg}8.9
            \\
            \rowcolor{TableGray3}
            (c)&
            \cellcolor{TableGray3}Late Fusion &
            \cellcolor{TableGray3}\no & 
            \cellcolor{TableGray3}ViT-Small &
            \cellcolor{colorslideyesleg}12.5 & \cellcolor{colorslideyesleg}6.7 &
            \cellcolor{colorslidenoleg}6.9 & \cellcolor{colorslidenoleg}4.5
            \\
            \specialrule{0.5pt}{0pt}{0pt}
            (d)&
            ChannelCat &
            \no & 
            ResNet9 &
            \cellcolor{colorslideyesleg}83.2 & \cellcolor{colorslideyesleg}43.9 & 
            \cellcolor{colorslidenoleg}44.6 & \cellcolor{colorslidenoleg}23.4
            \\
            (e)&
            ChannelCat \cite{sun2024fgprompt}& 
            \yes & 
            ResNet9 &
            \cellcolor{colorslideyesleg}83.6 & \cellcolor{colorslideyesleg}42.1 &
            \cellcolor{colorslidenoleg}31.7 & \cellcolor{colorslidenoleg}18.7
            \\
            (f)&
            ChannelCat &
            \no & 
            ViT-Small &
            \cellcolor{colorslideyesleg}71.1 & \cellcolor{colorslideyesleg}34.3 &
            \cellcolor{colorslidenoleg}35.3 & \cellcolor{colorslidenoleg}16.2
            \\
            (g)&
            Cross-attn &
            \no & 
            DEBiT-B &
            \cellcolor{colorslideyesleg}0.0 & \cellcolor{colorslideyesleg}0.0 &
            \cellcolor{colorslidenoleg}0.0 & \cellcolor{colorslidenoleg}0.0
            \\
            \rowcolor{colorpretrainleg}
            (h)&
            Cross-attn \cite{CrocoNav2024}&
            \no & 
            DEBiT-B$^\ddagger$ &
            \cellcolor{colorslideyesleg}\textbf{90.5} & \cellcolor{colorslideyesleg}\textbf{60.3} &
            \cellcolor{colorslidenoleg}\textbf{81.7} & \cellcolor{colorslidenoleg}\textbf{52.0}
            \\
            \specialrule{1pt}{0pt}{0pt}
    \end{tabular}
    \caption{\label{tab:main_result}\textbf{Agents with different visual encoders} trained and validated with \boxslideyes{\code{Sliding=True}} or \boxslideno{\code{Sliding=False}}. \textbf{s2d}$^\dagger$=\textit{SpaceToDepth}.\boxpretrain{$\ddagger$=pre-trained for RPVE}.
    }
}
\end{table}

\subsection{Benchmarking architectures}
\label{sec:benchmarkingarchitectures}
\noindent
\cref{tab:main_result} summarizes results for different architectures and settings. With \boxslideyes{\code{Sliding=True}}, \textit{ChannelCat} (d)-(f) obtains excellent performance, close to \textit{DEBiT-B} (h) which has a larger, more complex architecture, and is pre-trained on RPVE. Without pre-training, \textit{DEBiT} is not exploitable, as also reported in \cite{CrocoNav2024}. \textit{Late Fusion} architectures (a)-(c) underperform, and \textit{SpaceToDepth} has no significant impact on either \textit{ChannelCat} or \textit{Late Fusion} models 
: being able to compare local patches across images in a single layer through convolutions does not translate into numerical gains.
With \boxslideno{\code{Sliding=False}} the trends change dramatically. While the impact on the (previously already underperforming) \textit{Late Fusion} architecture is similar, \textit{ChannelCat}, (d)-(f), now breaks down and performance is halved, or less. In contrast, \textit{DEBiT} is able to cope well with the more realistic \code{Sliding=False} setting, arguably because of its strong pre-trained visual encoder, confirming the importance of visual pre-training. 

To summarize, coming back to \textit{Q1} and \textit{Q2} raised in \cref{sec:intro}, 
our conclusions are the following:
\begin{description}[nosep,,itemsep=1mm,labelindent=0mm,leftmargin=3mm,topsep=1mm]
\item[R1:] In realistic settings (\code{Sliding=True}), agents trained with RL alone perform significantly worse than \textit{DEBiT} with pre-trained visual encoder.
\item[R2:] \textit{Early Fusion} agents perform better than \textit{Late Fusion} counterparts, with early patch-wise fusion being the best option, but requiring pre-training.
\end{description}
Additionally, we found that recent success of agents based on small CNNs trained end-to-end with RL only, was mostly due to setting \code{Sliding=True} in Habitat.

\subsection{Transferring capabilities obtained w. ``Sliding''}
\noindent
The huge impact of the \code{Sliding} simulator setting came somewhat as a surprise. While its role in sim2real transfer of point goal navigation was established in \cite{kadian20sm2real}, the common conjecture was that it impacted ``low-level navigation'', \ie, finding free space, avoiding obstacles, capabilities which are independent of the extraction of goal directional information. Our experiments, however, suggest that the choice of binocular visual encoder $\phi$ is impacted by this hyper-parameter, which 
should be related to motion only.

This then raises the question, whether (i) the two agents trained in their respective settings have similar capabilities but perform differently purely due to the difference in task difficulty, or (ii) whether, for some reason, training with sliding actually leads to different and perhaps better performing agent capabilities. We tested these hypotheses by 
performing experiments loading the weights of an agent trained with \code{Sliding=True} and finetuning it 
with \code{Sliding=False}. For these experiments we used the \textit{ChannelCat + SpaceToDepth} agent, for which the difference in SR dropped from 83.6\% to 31.7\% when sliding was switched off (cf.  \cref{tab:main_result}).

\cref{tab:slidingtransfer} shows the results of these experiments. Taking the agent trained with \code{Sliding=True} and validating it on \code{False}, row (b), makes SR jump to 54.6\%, compared to the baseline of 31.7\%, trained on \code{False} only. This is a highly surprising result. Although the performance is a far cry of the 83.6\% obtained by training and validating with \code{True}, it indicates that a certain amount of transfer is possible --- some capabilities are learned with sliding enabled. Finetuning this agent for 100M steps on \code{False} provides further gains and yields SR=65.7\%, row (c).

\begin{table}[t] \centering
{\small
    \setlength{\tabcolsep}{1pt}
    \begin{tabular}{llcccc|cc}
            \specialrule{1pt}{0pt}{0pt}
            \rowcolor{TableGray2}      
             &&
             {\scriptsize \textbf{Perception}} & 
             \multicolumn{3}{c|}{\scriptsize{\textbf{Action}}} &
             \multicolumn{2}{c}{\textbf{(\%)}} 
             \\
             \rowcolor{TableGray2}      
             & \multirow{-2}{*}{\textbf{Checkpoint}} &
             $\phi$ & 
             $h$ &
             $\zeta$ &
             $\pi$ &
             \textbf{SR} & 
             \textbf{SPL}
             \\
            \specialrule{0.5pt}{0pt}{0pt}   
            \cellcolor{TableGray2} (a) & 
            \cellcolor{TableGray2} Load all ``false'' &
            \cellcolor{colfreeze} \wfalse\wfr &
            \cellcolor{colfreeze} 
            \wfalse\wfr &
            \cellcolor{colfreeze} 
            \wfalse\wfr &
            \cellcolor{colfreeze} 
            \wfalse\wfr &
            31.7 & 18.7 
            \\ 
            \cellcolor{TableGray2} (b) & 
            \cellcolor{TableGray2} Load all ``true'' &
            \cellcolor{colfreeze} \wtrue\wfr &
            \cellcolor{colfreeze} 
            \wtrue\wfr &
            \cellcolor{colfreeze} 
            \wtrue\wfr &
            \cellcolor{colfreeze} 
            \wtrue\wfr &
            54.6 & 27.5 \\  
            \cellcolor{TableGray2} (c) & 
            \cellcolor{TableGray2} Load all ``true''&
            \cellcolor{colfinetune} 
            \wtrue\wft &
            \cellcolor{colfinetune} 
            \wtrue\wft &
            \cellcolor{colfinetune} 
            \wtrue\wft &
            \cellcolor{colfinetune} 
            \wtrue\wft &
            \textbf{65.7} & \textbf{34.1}\\
            \cellcolor{TableGray2} (d) & 
            \cellcolor{TableGray2} Load action ``true''&
            \cellcolor{coltrain} $\circlearrowright$ &
            \cellcolor{colfreeze} 
            \wtrue\wfr &
            \cellcolor{colfreeze} 
            \wtrue\wfr &
            \cellcolor{colfreeze} 
            \wtrue\wfr &
            ~0.0 & ~0.0
            \\    
            \cellcolor{TableGray2} (e) & 
            \cellcolor{TableGray2} Load action ``true'' &
            \cellcolor{coltrain} 
            \wscratch &
            \cellcolor{colfinetune} 
            \wtrue\wft &
            \cellcolor{colfinetune} 
            \wtrue\wft &
            \cellcolor{colfinetune} 
            \wtrue\wft &
            ~6.1 & ~4.8 
            \\
            \cellcolor{TableGray2} (f) & 
            \cellcolor{TableGray2} Load perception ''true'' &
            \cellcolor{colfreeze} 
            \wtrue\wfr &
            \cellcolor{coltrain} 
            \wscratch &
            \cellcolor{coltrain} 
            \wscratch &
            \cellcolor{coltrain} 
            \wscratch &
            26.4 & 14.3
            \\
            \cellcolor{TableGray2} (g) & 
            \cellcolor{TableGray2} Load perception ``true''&
            \cellcolor{colfinetune} 
            \wtrue\wft &
            \cellcolor{coltrain} 
            \wscratch &
            \cellcolor{coltrain} 
            \wscratch &
            \cellcolor{coltrain} 
            \wscratch &
            38.5 & 20.3 
            \\            
            \specialrule{1pt}{0pt}{0pt}
    \end{tabular}
\caption{\label{tab:slidingtransfer}\textbf{OOD behavior and cross-domain transfer (\textit{ChannelCat+SpaceToDepth agent}):} Can navigation knowledge learned with sliding be transferred to the non-sliding setting through fine-tuning? \wfalse: load from agent trained with \code{Sliding=False},
\wtrue: load from agent trained with \code{Sliding=True},
\boxfreeze{\wfr: frozen},
\boxfinetune{\wft: finetune}
, \boxtrain{\wscratch: re-train from scratch}.}
}
\end{table}

To pinpoint the effect, we loaded different parts of the agent trained with \code{Sliding=True}: only the perception part (weights of the visual encoder $\phi$), only the action part (GRU $h$, policy $\pi$, previous action encoder $\zeta$) or both. 
Transferring the action part of the agent, \ie GRU $h$, policy $\pi$, encoder $\zeta$, rows (d) and (e), does not lead to any discernible performance: learning the upstream visual encoder $\phi$ from scratch through pre-loaded downstream weights of the GRU and policy seems a very hard task. However, transferring the perception weights $\phi$ and training the action part from scratch, rows (f) and (g), leads to exploitable results, although significantly lower than loading the full agent weights. With a SR of 38.5\% for the finetuning version, row (g), these are still higher than the baseline of 31.7\% trained from scratch on \code{False}. In \cref{tab:slidingtransferdebit} we perform similar experiments with \textit{DEBiT} which, however, uses a pre-trained visual encoder $\phi$ that is kept frozen. We load the rest of the agent trained with \code{Sliding=True} and evaluate it with \code{False} (b): in this case, no transfer happens and performance drops significantly. Finetuning the agent trained with \code{True} on \code{False}, (c), improves the \code{Sliding=True} policy, but not to the extent of reaching the performance of the agent trained on-distribution, \ie~\code{Sliding=False}, arguably because RPVE pre-training of perception is already providing a strong visual encoder.

\begin{table}[t] \centering
{\small
    \setlength{\tabcolsep}{1pt}
    \begin{tabular}{llcccc|cc}
            \specialrule{1pt}{0pt}{0pt}
            \rowcolor{TableGray2}      
             &&
             {\scriptsize \textbf{Perception}} & 
             \multicolumn{3}{c|}{\scriptsize{\textbf{Action}}} &
             \multicolumn{2}{c}{\textbf{(\%)}} 
             \\
             \rowcolor{TableGray2}      
             & \multirow{-2}{*}{\textbf{Checkpoint}} &
             $\phi$ & 
             $h$ &
             $\zeta$ &
             $\pi$ &
             \textbf{SR} & 
             \textbf{SPL}
             \\
            \specialrule{0.5pt}{0pt}{0pt}   
            \cellcolor{TableGray2} (a) &
            \cellcolor{TableGray2} Load all ``false'' &
            \cellcolor{colfreeze} 
            \wpt\wfr &
            \cellcolor{colfreeze} 
            \wfalse\wfr &
            \cellcolor{colfreeze} 
            \wfalse\wfr &
            \cellcolor{colfreeze} 
            \wfalse\wfr &
            \textbf{81.7} & \textbf{52.0} 
            \\ 
            \cellcolor{TableGray2} (b) &
            \cellcolor{TableGray2} Load all ``true'' &
            \cellcolor{colfreeze} 
            \wpt\wfr &
            \cellcolor{colfreeze} 
            \wtrue\wfr &
            \cellcolor{colfreeze} 
            \wtrue\wfr &
            \cellcolor{colfreeze} 
            \wtrue\wfr &
            59.5 & 35.4 \\
            \cellcolor{TableGray2} (c) &
            \cellcolor{TableGray2} Load all ``true'' &
            \cellcolor{colfreeze} 
            \wpt\wfr &
            \cellcolor{colfinetune} 
            \wtrue\wft &
            \cellcolor{colfinetune} 
            \wtrue\wft &
            \cellcolor{colfinetune} 
            \wtrue\wft &
            79.6 & 46.9 \\
            \specialrule{1pt}{0pt}{0pt}
    \end{tabular}
\caption{\label{tab:slidingtransferdebit}\textbf{OOD behavior (\textit{DEBiT agent}):} \wfalse: load from agent trained with \code{Sliding=False},
\wtrue: load from agent trained with 
\code{Sliding=True}, \wpt: pre-trained with RPVE,
\boxfreeze{\wfr: frozen}, \boxfinetune{\wft: finetune}.}
}
\end{table}

We draw the following lessons from these experiments. The difference between  \code{Sliding=True} or \code{False} cannot only be described by the task difficulty alone. The easier task (\code{True}) allows to learn additional capabilities, which transfer to the harder task, and which are partially related to perception (since the performance in \cref{tab:slidingtransfer}(g) $>$ \ref{tab:slidingtransfer}(a)), and also related to action / sequential decision taking, since \cref{tab:slidingtransfer}(b) $\gg$ \ref{tab:slidingtransfer}(a) and \cref{tab:slidingtransfer}(b) $\gg$ \ref{tab:slidingtransfer}(g). However, we conjecture that the transfer of the knowledge stored in the action component is \textit{also} highly linked to improving the perception skills, as it does not happen when $\phi$ is pre-trained and frozen (\cref{tab:slidingtransferdebit}(a) $\gg$ \ref{tab:slidingtransferdebit}(b), and \cref{tab:slidingtransferdebit}(a)~$>$~\ref{tab:slidingtransferdebit}(c)). One possible explanation is that only training with \code{False} leads to undertraining of, both, action and perception: the policy keeps getting stuck (which we empirically confirmed) and does not learn to cope with the last meters of each episode; this, in turn, leads to undertraining the comparison between the (hardly ever seen) goals and observations.

\subsection{Probing RPVE capabilities}
\noindent
Given that a certain limited capacity for visual reasoning has been learned by the visual encoder $\phi$ when the full agent has been trained for navigation with \code{Sliding=True}, we pursue this question further and directly investigate how well the different visual encoders can deal with extracting directional information. Or, motivated differently, given the potential impact on relative pose estimation of training these visual encoders from navigation losses only without pose supervision, we evaluate how well they work for this task. We took the frozen visual encoders $\phi$ of the agents evaluated in \cref{tab:main_result} and trained probing heads $p$ on top of them predicting relative pose and visibility (RPVE):
\setlength{\abovedisplayskip}{3pt}
\setlength{\belowdisplayskip}{3pt}
\begin{equation}
(\mathbf{t},\mathbf{R},v) = p\Big(\phi(\mathbf{o},\mathbf{g})\Big).
\label{eq:probing}
\end{equation}
In line with \cite{CrocoNav2024}, relative pose between the observed image and the goal image is composed of two components, translation $\mathbf{t}{\in}\mathbb{R}^3$ and a rotation matrix $\mathbf{R}{\in}\mathbb{R}^{3\times 3}$. Visibility quantifies the amount of overlap between the two images, necessary since in navigation settings the goal image might not even be observed at certain moments. It is defined  as the proportion of $16{\times}16$ patches of the goal image $\mathbf{g}$ which are visible in the observed image $\mathbf{o}$.

\begin{table}[t] \centering
{\small
    \setlength{\tabcolsep}{1pt}
    \begin{tabular}{lcccc|ccc}
            \specialrule{1pt}{0pt}{0pt}
            \rowcolor{TableGray2} 
             \multicolumn{2}{c}{\textbf{Model}} &
              &
              & 
              &
             \multicolumn{2}{c}{\scriptsize  \textbf{\%corr.poses}} &
             {\scriptsize  \textbf{\%corr.vis.}}
             \\
             \rowcolor{TableGray2}
             \multicolumn{2}{c}{(Table nr. + row)}& 
             \multirow{-2}{*}{\textbf{s2d$^\dagger$}}
             &
             \multirow{-2}{*}{\textbf{Backbone}}
             &
             \multirow{-2}{*}{\textbf{S}$^\ddagger$}
             &
             {\scriptsize \textbf{1m,10\textdegree}} & 
             {\scriptsize \textbf{2m,20\textdegree}} & {\scriptsize$\mathbf{{<}0.05}$}
             \\
            \specialrule{0.5pt}{0pt}{0pt}
            \rowcolor{colorslideyes}
            \cellcolor{colorslideyesleg} Late Fusion &
            \cellcolor{colorslideyesleg} \ref{tab:main_result}a &
            \cellcolor{colorslideyesleg} \no & 
            \cellcolor{colorslideyesleg} ResNet9 & 
            \cellcolor{colorslideyesleg} \yes & 
            ~7.6 & 26.0 & 13.8 \\
            \rowcolor{colorslideyes}
            \cellcolor{colorslideyesleg} Late Fusion &
            \cellcolor{colorslideyesleg} \ref{tab:main_result}b&
            \cellcolor{colorslideyesleg} \yes & 
            \cellcolor{colorslideyesleg} ResNet9 & 
            \cellcolor{colorslideyesleg} \yes & 
            ~9.0 & 29.6 & 16.1
            \\
            \rowcolor{colorslideyes}
            \cellcolor{colorslideyesleg} ChannelCat &
            \cellcolor{colorslideyesleg} \ref{tab:main_result}d&
            \cellcolor{colorslideyesleg} \no & 
            \cellcolor{colorslideyesleg} ResNet9 & 
            \cellcolor{colorslideyesleg} \yes & 
            11.4 & 29.3 & 13.7 
            \\
            \rowcolor{colorslideyes}
            \cellcolor{colorslideyesleg} ChannelCat & 
            \cellcolor{colorslideyesleg} \ref{tab:main_result}e&
            \cellcolor{colorslideyesleg} \yes & 
            \cellcolor{colorslideyesleg} ResNet9 & 
            \cellcolor{colorslideyesleg} \yes & 
            18.4 & 41.6 & 20.8
            \\            
            \specialrule{0.5pt}{0pt}{0pt}
            \rowcolor{colorslideno}
            \cellcolor{colorslidenoleg} Late Fusion &
            \cellcolor{colorslidenoleg} \ref{tab:main_result}a&
            \cellcolor{colorslidenoleg} \no & 
            \cellcolor{colorslidenoleg} ResNet9 & 
            \cellcolor{colorslidenoleg} \no & 
            ~7.8 & 26.8 & 13.2 
            \\
            \rowcolor{colorslideno}
            \cellcolor{colorslidenoleg} Late Fusion &
            \cellcolor{colorslidenoleg} \ref{tab:main_result}b &
            \cellcolor{colorslidenoleg} \yes & 
            \cellcolor{colorslidenoleg} ResNet9 & 
            \cellcolor{colorslidenoleg} \no & 
            ~8.7 & 28.5 & 16.1
            \\
            \rowcolor{colorslideno}
            \cellcolor{colorslidenoleg} ChannelCat &
            \cellcolor{colorslidenoleg} \ref{tab:main_result}d &
            \cellcolor{colorslidenoleg} \no & 
            \cellcolor{colorslidenoleg} ResNet9 & 
            \cellcolor{colorslidenoleg} \no & 
            ~9.8 & 26.9 & 13.8 
            \\
            \rowcolor{colorslideno}
            \cellcolor{colorslidenoleg} ChannelCat & 
            \cellcolor{colorslidenoleg} \ref{tab:main_result}e &
            \cellcolor{colorslidenoleg} \yes & 
            \cellcolor{colorslidenoleg} ResNet9 & 
            \cellcolor{colorslidenoleg} \no & 
            12.5 & 31.9 & 19.2            
            \\
            \specialrule{0.5pt}{0pt}{0pt}
            \rowcolor{colorfinetune}
            \cellcolor{colorfinetuneleg} ChannelCat & 
            \cellcolor{colorfinetuneleg} \ref{tab:slidingtransfer}c &
            \cellcolor{colorfinetuneleg} \yes & 
            \cellcolor{colorfinetuneleg} ResNet9 & 
            \cellcolor{colorfinetuneleg} \wft & 
            18.2 & 41.4 & 21.1 
            \\
            \rowcolor{colorfinetune}
            \cellcolor{colorfinetuneleg} ChannelCat & 
            \cellcolor{colorfinetuneleg} \ref{tab:slidingtransfer}d &
            \cellcolor{colorfinetuneleg} \yes & 
            \cellcolor{colorfinetuneleg} ResNet9 & 
            \cellcolor{colorfinetuneleg} \wft & 
            ~5.8 & 22.9 & ~6.7
            \\
            \rowcolor{colorfinetune}
            \cellcolor{colorfinetuneleg} ChannelCat & 
            \cellcolor{colorfinetuneleg} \ref{tab:slidingtransfer}e &
            \cellcolor{colorfinetuneleg} \yes & 
            \cellcolor{colorfinetuneleg} ResNet9 & 
            \cellcolor{colorfinetuneleg} \wft & 
            ~7.2 & 26.1 & 11.9
            \\
            \rowcolor{colorfinetune}
            \cellcolor{colorfinetuneleg} ChannelCat & 
            \cellcolor{colorfinetuneleg} \ref{tab:slidingtransfer}g &
            \cellcolor{colorfinetuneleg} \yes & 
            \cellcolor{colorfinetuneleg} ResNet9 & 
            \cellcolor{colorfinetuneleg} \wft & 
            18.6 & 41.6 & 21.0
            \\
            \specialrule{0.5pt}{0pt}{0pt}
            \rowcolor{colorpretrain}
            \cellcolor{colorpretrainleg} \textcolor{gray}{\textit{Cross-att}} &
            \cellcolor{colorpretrainleg} \textcolor{gray}{\ref{tab:main_result}h}&
            \cellcolor{colorpretrainleg} \no & 
            \cellcolor{colorpretrainleg} \textcolor{gray}{DEBiT-B} & 
            \cellcolor{colorpretrainleg} \textcolor{gray}{N/A} & 
            \textcolor{gray}{92.1} & \textcolor{gray}{96.8} & \textcolor{gray}{88.8}
            \\
            \rowcolor{colorpretrain}
            \cellcolor{colorpretrainleg} \textcolor{gray}{\textit{Late fusion}} &
            \cellcolor{colorpretrainleg} &
            \cellcolor{colorpretrainleg} \no & 
            \cellcolor{colorpretrainleg} \textcolor{gray}{DEBiT-B} & 
            \cellcolor{colorpretrainleg} \textcolor{gray}{N/A} & 
            \textcolor{gray}{14.8} & \textcolor{gray}{38.6} & \textcolor{gray}{19.6} 
            \\
            \rowcolor{colorpretrain}
            \cellcolor{colorpretrainleg} \textcolor{gray}{\textit{Late fusion}} &
            \cellcolor{colorpretrainleg} &
            \cellcolor{colorpretrainleg} \no & 
            \cellcolor{colorpretrainleg} \textcolor{gray}{DINOv2} & 
            \cellcolor{colorpretrainleg} \textcolor{gray}{N/A} & 
            \textcolor{gray}{12.9} & \textcolor{gray}{34.0} & \textcolor{gray}{22.7} 
            \\
            \specialrule{1pt}{0pt}{0pt}
    \end{tabular}
\caption{\label{tab:probing} \textbf{Probing relative pose and visibility estimation:} representations $\phi(\mathbf{o},\mathbf{g})$ trained with an RL (navigation) loss are frozen, and then we train a probing head. \boxpretrain{\textcolor{gray}{\textit{The last block of methods in italic}}} is not comparable, as the encoders were pre-trained: \textit{DEBiT} \cite{CrocoNav2024} was pre-trained on RPVE losses, DINOv2 \cite{oquab2023dinov2} with SSL. We report RPVE performance on a hold-out set. \textbf{s2d}$^\dagger$=\textit{SpaceToDepth}, \textbf{S}$^\ddagger$ = \code{Sliding=True}. The \boxtransfer{third block} shows agents which have been finetuned (\wft) from \code{True} to \code{False}, cf. \cref{tab:slidingtransfer}.}
}
\end{table}

The probing heads $p$ are adapted to the different architectures, while at the same time providing a comparable capacity of $\sim{3M}$ parameters for each head variant. They collect features at the last spatial representation (feature map for ResNet, patch embeddings for ViT, concatenation of these for \textit{Late Fusion} representations) and first linearly project token-wise to a lower dimension, before flattening to obtain a global representation which is passed to an MLP with 1024 hidden units. The low dimension of the projection changes according to architectures (as the spatial resolution, \ie, the number of tokens change) to make the parameter counts similar: 64 for ViTs, 192 (resp. 16) for ResNet9 with (resp. without) \textit{SpaceToDepth}.


As in~\cite{CrocoNav2024}, we generated a probing dataset by combining the 3D scene datasets Gibson~\cite{xiazamirhe2018gibsonenv}, MP3D~\cite{chang2018matterport3d}, and HM3D~\cite{ramakrishnan2021hm3d}, following their standard train/val scene splits. We sample pairs of points randomly in the scene, calculate the shortest path between them, and extract 10 intermediate poses on the path. These 10 poses are then each combined with the end point to form pairs of observation $\mathbf{o}$ and goal $\mathbf{g}$. Ground-truth for relative pose is directly available from the simulator, and visibility can be estimated from the two point clouds generated from the respective images. We generated around 68M image pairs total.

We train the probing heads $p$ with a loss combining all three components, translation, rotation and visibility, where pose supervision is switched off when visibility is low,
\begin{align}
\mathcal{L}_{p} = \sum\mathop{}_{\mkern-5mu i}
  \Bigl[
    |v_i {-} v_i^*| + \mathbf{1}_{v^*_i{>}\tau}
    \bigl\{
    |\mathbf{t}_i {-} \mathbf{t}_i^*| + |\mathbf{R}_i {-} \mathbf{R}_i^*|
    \bigr\}
  \Bigr] ,
\label{eq:loss_full}
\end{align}
where $i$ indexes image pairs over the probing dataset, $\mathbf{t}_i^*, \mathbf{R}_i^*, v_i^*$ denote ground truth values, $\mathbf{1_.}$ is the binary indicator function, $|.|$ denotes the $L_1$ loss and $\tau$ is a threshold.

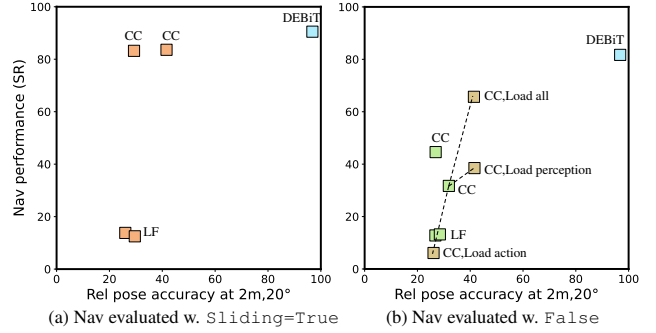
\begin{figure} \centering
    \begin{minipage}{0.51\linewidth}
    \begin{tikzpicture}
        \draw (0, 0) node[anchor=west,inner sep=0] {
        \includegraphics[width=\linewidth]{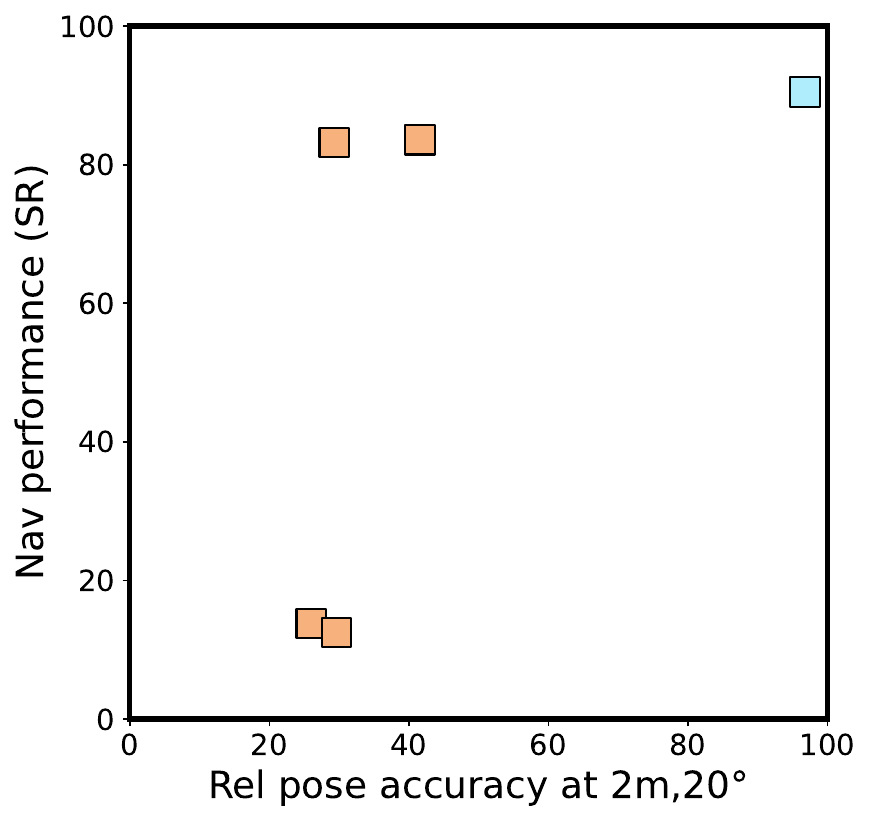}};
        \draw (3.6, 1.8) node[anchor=west,inner sep=0] {{\tiny DEBiT}};   
        \draw (1.75, -1.05) node[anchor=west,inner sep=0] {{\tiny LF}};   
        \draw (2, 1.55) node[anchor=west,inner sep=0] {{\tiny CC}};   
        \draw (1.5, 1.55) node[anchor=west,inner sep=0] {{\tiny CC}}; 
        \end{tikzpicture}        
    \end{minipage}
    \begin{minipage}{0.48\linewidth}        
    \begin{tikzpicture}
        \draw (0, 0) node[anchor=west,inner sep=0] {
        \includegraphics[width=\linewidth]{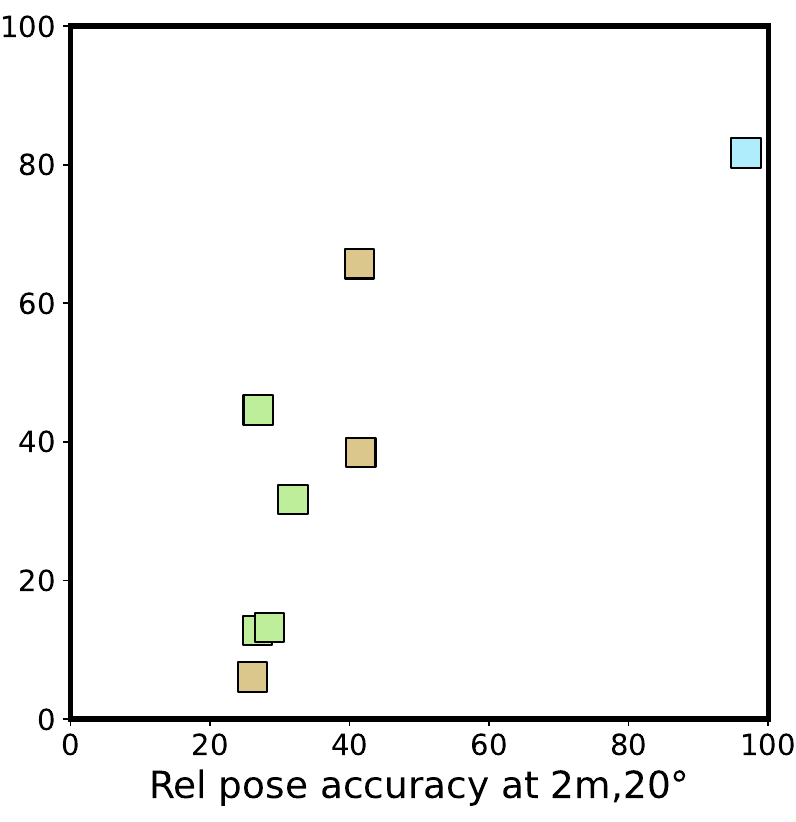}};
        \draw (3.3, 1.5) node[anchor=west,inner sep=0] {{\tiny DEBiT}};   
        \draw (1.95, 0.75) node[anchor=west,inner sep=0] {{\tiny CC,Load all}};   
        \draw (1.95, -0.25) node[anchor=west,inner sep=0] {{\tiny CC,Load perception}}; 
        \draw (1.25, 0.2) node[anchor=west,inner sep=0] {{\tiny CC}};   
        \draw (1.6, -0.5) node[anchor=west,inner sep=0] {{\tiny CC}};   
        \draw (1.5, -1.1) node[anchor=west,inner sep=0] {{\tiny LF}};   
        \draw (1.4, -1.35) node[anchor=west,inner sep=0] {{\tiny CC,Load action}};
        \draw[dash pattern=on 1.5pt off 1pt] (1.48,-0.45) -- (1.8,0.75);
        \draw[dash pattern=on 1.5pt off 1pt] (1.48,-0.45) -- (1.83,-0.2);
        \draw[dash pattern=on 1.5pt off 1pt] (1.48,-0.45) -- (1.265,-1.35);
    \end{tikzpicture}
    \end{minipage} 
        \hspace{0.1cm}
        {\scriptsize 
        (a) Nav evaluated w. \code{Sliding=True} \hspace{0.5cm} 
        (b) Nav evaluated w. \code{False}}  
    \caption{\label{fig:probe_vs_nav}\textbf{Nav \vs Rel-pose:} navigation perf. (SR,\%) plotted against pose est. probing accuracy (\% for err {\footnotesize ${<}2m,20^\circ$}) for 4 types of visual encoders $\phi$: 
    \boxslideyes{trained w. sliding}, 
    \boxslideno{trained w/o sliding},
    \boxpretrain{pre-trained w. RPVE},
    \boxtransfer{trained w. sliding, finetuned w/o}; (LF = Late~Fusion, CC=ChannelCat). The dashed line \sampleline{dash pattern=on 1.5pt off 1pt} relates the \boxtransfer{finetuned models} to the \boxslideno{same model  trained w/o sliding}.
    }
\end{figure}
\begin{figure*}[t] \centering
\includegraphics[width=1\linewidth]{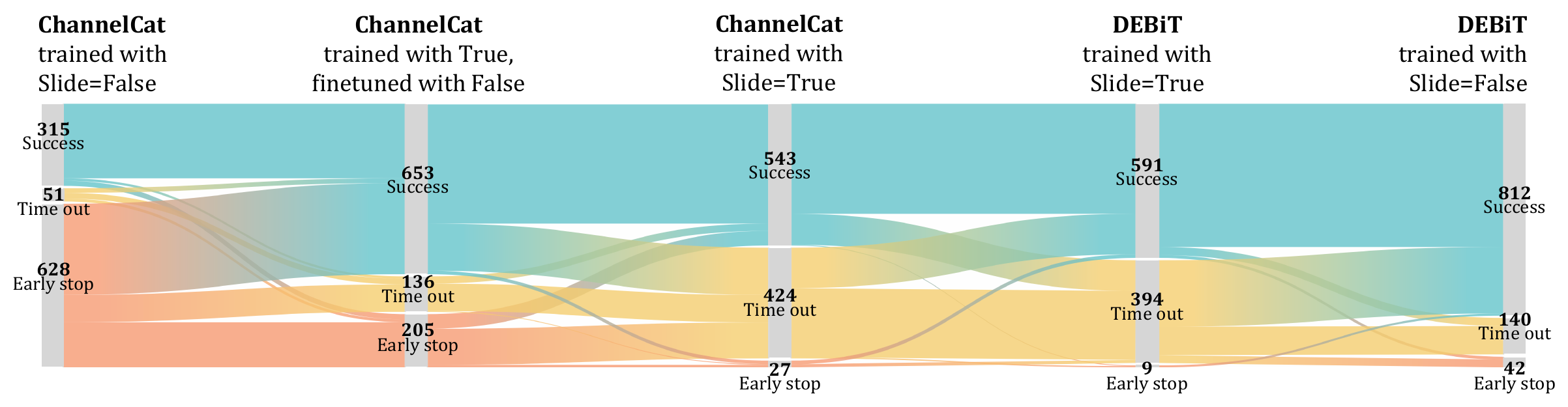} \\[-0.1cm]
\caption{\label{fig:sankey_plots}\textbf{Analysis of navigation behavior:} Sankey plots show the distribution of success/failure codes over 994 test episodes for different models, and their ``flow'' between certain pairs of models. For instance, the strength of the connection between ``Time out'' (left) and ``success'' (right) indicates how many episodes toggled from one to the other when switching from the left to the right model.
}
\end{figure*}

\myparagraph{Metrics} relative pose is evaluated over the pairs with visibility over $\tau$ in the percentage of correct poses for given thresholds on distance and angle, \eg 1 meter and 10\textdegree.  Visibility is evaluated over all pairs by its accuracy at $\pm$0.05, \ie, the percentage of prediction within a 0.05 margin of the ground-truth value.

\myparagraph{Results} \cref{tab:probing} gives results comparing RPVE performance on a validation set for all tested visual encoders, and \cref{fig:probe_vs_nav} provides a scatter plot relating it to navigation performance (taken from \cref{tab:main_result}, \ref{tab:slidingtransfer}).  While the \textit{ChannelCat} architecture obtains high navigation performance in the \code{Sliding=True} setting, \ref{fig:probe_vs_nav}(a), its pose estimation performance remains limited. \textit{Late Fusion} models generally provide low performance both for navigation and pose estimation. In the \code{Sliding=False} setting, \ref{fig:probe_vs_nav}(a), the navigation performance significantly drops for the \textit{ChannelCat} models, as already discussed in \cref{sec:benchmarkingarchitectures}, and also somewhat for pose estimation. Transferring the models trained with \code{True} on the \code{False} setting provides gains not only in navigation performance, as already discussed, but also on pose estimation --- only when at least the weights of the visual encoder $\phi$ are transferred, but in particular when the whole agent is transferred and finetuned. This further corroborates our conjecture, that the perception model is undertrained when sliding is disabled. Finally, the best performance is obtained by the \textit{DEBiT} model based on ViTs and cross-attention, both in navigation and pose probing. However, the pose estimation performance is \textit{not comparable for obvious reasons} --- the model has been pre-trained and frozen for this  task. Its performance is given for information.

In \cref{tab:probing} we also provide pose probing results for two additional models, where we explore whether the pose estimation performance of \textit{Late Fusion} can be increased by using two different pre-trained and frozen encoders, namely DINOv2 and the ViT encoder part of \textit{DEBiT}, without the cross-attention layers combining them\footnote{\textit{DEBiT} first encodes each image with a ViT and then passes both embeddings to a decoder using cross-attention, cf. \cite{CrocoNav2024}.}. They both underperform, further corroborating the importance of comparing visual representations early, allowing to compute correspondences between local image parts. These experiments support a positive answer to \textit{Q3} in \cref{sec:intro}: there is a clear correlation between navigation and RPE performance.

\subsection{Analysis of navigation behavior}
\noindent
In \cref{fig:sankey_plots} we provide Sankey plots showing the differences in navigation behavior as a distribution of success/error types over 994 test episodes evaluated in environments with \code{Sliding=False}, and how this behavior varies across different models and different training settings. Here we distinguish between unsuccessful episodes due to \textit{Time~out}, when the agent does not complete an episode in the 1000 steps budget, and \textit{Early~stop}, when the agent terminates the episode but is further away than the 1m success threshold. 

Left, the \textit{ChannelCat} agent trained on \code{Sliding=False}, \cref{tab:slidingtransfer}(a), stops too early in most of the episodes, confirming that its components are not properly trained, probably due to the difficulty of the task. Next, finetuning on \code{False} the  \textit{ChannelCat} agent trained with \code{True}, \cref{tab:slidingtransfer}(c), allows to obtain an agent with good navigation capabilities. It improves the \textit{ChannelCat} agent trained with \code{True}, \cref{tab:slidingtransfer}(b), in the middle in \cref{fig:sankey_plots}, whose main failure mode is time out. We observe that most of these time-outs occur far from the goal and are due to the agent getting stuck, likely because it learned to slide along obstacles, which is not possible with \code{False}. On the 4th column, \textit{DEBiT} trained with \code{True}, \cref{tab:slidingtransferdebit}(b), exhibits a similar behavior, while \textit{DEBiT} trained with \code{False}, \cref{tab:slidingtransferdebit}(a), on the right-most column, can effectively solve the complex navigation task, arguably thanks to its pretrained, high capacity, perception module.

\section{Discussion and conclusion}
\noindent
We have studied image goal navigation and raised questions on agent architecture and whether components dedicated to vision or decision taking should be trained jointly with RL or separately. We have shown that the success of recent architectures based on channel stacking can be linked to a simulator setting allowing sliding. While agents trained with sliding enabled are known to transfer badly to real environments, where this is not possible, our analysis has shown that capabilities learned with this sliding setting can actually be partially transferred to more realistic environment settings. While sliding is a property of agent dynamics and not inherently linked to perception, we have shown that it significantly impacts training of the visual encoders of end-to-end trained agents: \textit{transferring capabilities to the realistic setting only seems to be successful if it also involves transferring the visual encoder weights}. We conjecture that training with the realistic setting leads to undertraining both the perception module, as the goal is seen more rarely, and the rest of the agent, which has a difficulty to cover later portions of the episodes. 

We also compared and analyzed different widely-used architectures: \textit{Late Fusion, ChannelCat, SpaceToDepth} projections and \textit{Cross-attention}. Results suggest that \textit{successful architectures require support for early fusion of representations}, allowing to compute correspondences between local parts of images: \textit{Late Fusion} generally underperforms. As expected, we also found a \textit{correlation between navigation performance and (emerging) relative pose estimation performance}, which we probed with explicit heads. Finally, we argue that, 
up to our knowledge, there does not seem to exist a simple solution to learn with RL alone image goal navigation 
end-to-end with simple 
low-capacity architectures and without pre-training, which we judge to remain essential for goal-oriented navigation.

{\small
\bibliographystyle{ieee_fullname}
\bibliography{main}
}

\clearpage
\appendix

\section{Agents implementation details}

\myparagraph{ResNet backbone} The ResNet implementation is borrowed from~\cite{ridnik2021tresnet,al2022zero,sun2024fgprompt} and the code is publicly available.\footnote{\url{https://github.com/XinyuSun/FGPrompt}} The default network is implemented as a custom ResNet9 with \textit{SpaceToDepth} blocks~\cite{ridnik2021tresnet} that rearrange the input of each 2D convolution layer such that each of its spatial dimensions is divided by a factor 4, while the channel dimension is 16 times larger. In the paper we use the ResNet9 version because of the performance reported in~\cite{sun2024fgprompt}, in line with those of a ResNet50, but with a fraction of the parameters. In the ResNet version without \textit{SpaceToDepth}, these reshaping blocks are simply skipped. As customary, the 2D CNN output is flatten and passed to a linear layer to generate a 128D feature vector.

The \textit{Late Fusion} ResNet backbone uses two such networks and concatenates the output vectors before feeding them to the GRU, while in the \textit{ChannelCat} version the first 2D convolution layer is modified to receive as input a 6-channel image generated by channel-stacking observation and goal. 

\myparagraph{ViT backbone} We use the \textit{small} version of a popular ViT implementation,\footnote{\url{https://github.com/huggingface/pytorch-image-models}} augmented with the \textit{convolutional compression layer} proposed in \textit{OVRL-v2}~\cite{yadav2023ovrlv2} (code available in the Appendix of the \textit{OVRL-v2} paper) to compress the ViT output to a 2058D feature. We adopt this layer since \cite{yadav2023ovrlv2} reports significant improvements in navigation performance using it, over more common compression options like average pooling or [\code{CLS}] token. As discussed in~\cite{yadav2023ovrlv2}, without pre-training (as is the case here), ViT-Small performs better than ViT-Base despite having a fraction of the parameters, and therefore we use ViT-Small for all our experiments in the paper.

As for the ResNet backbone, the \textit{Late Fusion} network is implemented with two ViTs whose outputs are concatenated, while the \textit{ChannelCat} version has a modified input 2D convolution (first standard processing layer of the ViT) such that it can receive a 6-channel image composed by channel-stacking observation and goal. 

\myparagraph{\textit{DEBiT} backbone} In the paper we use the \textit{DEBiT-B} version of the architecture, the variant of \textit{DEBiT} with the best performance-size trade-off and publicly available with its pre-trained binocular encoder.\footnote{\url{https://github.com/naver/debit}} The binocular encoder is kept frozen and is based on the Geometric Foundation Model \emph{CroCo}~\cite{CroCo2022}, finetuned to estimate relative pose and visibility as described in~\cite{CrocoNav2024} and the main text. The network is implemented with a Siamese encoder, applied to observation and goal images, and a decoder which combines the output of these two encoders. Both encoders and decoders are implemented with a ViT architecture with self-attention layers, with the decoder also featuring cross-attention. The output of the decoder is further compressed by a fully connected layer (also kept frozen, from the \textit{DEBiT} model) that projects the flattened output of the decoder into a 3136D feature vector.

The \textit{DEBiT} model also features an observation encoder, needed since the binocular encoder is frozen and can thus only perform observation-goal comparison, without fulfilling other tasks needed for navigation such as, for example, free space estimation. It is implemented with a half-width ResNet18~\cite{he2016deep} that generates a 512D output feature.

We use the same \textit{DEBiT} architecture for the variant without pre-trained binocular encoder, the only difference being that the binocular encoder is trained from scratch with the other components of the agent using only RL. Similarly to what is reported in~\cite{CrocoNav2024}, we were not able to train such large architecture to a usable state using RL alone.

\begin{table*}[t] \centering
{\small
    \setlength{\tabcolsep}{2pt}
    \begin{tabular}{ll}
            \specialrule{1pt}{0pt}{0pt}
            \rowcolor{TableGray2}      
             \textbf{Model} & 
             \textit{ZSEL} \cite{al2022zero}\\
            \specialrule{0.5pt}{0pt}{0pt}   
            \textbf{Repository } & \scriptsize{\url{https://github.com/ziadalh/zero_experience_required}} \\
            \textbf{Configs} & \scriptcode{config/imagenav/eval\_ppo\_imagenav\_rgb.yaml}\\
            & $\hookrightarrow$ \scriptcode{config/imagenav/gibson/imagenav\_rgb.yaml}\\
            \textbf{Link} &
            \href{https://github.com/ziadalh/zero_experience_required/blob/dfe43e8ff9da74f27fe4db9ef311d9d40aeed67a/config/imagenav/gibson/imagenav_rgb.yaml#L15}{\scriptcode{github.com/ziadalh/zero\_experience\_required/blob/main/config/imagenav/gibson/imagenav\_rgb.yaml\#L15}}\\
            \specialrule{0.5pt}{0pt}{0pt}
            \rowcolor{TableGray2}      
             \textbf{Model} & 
             \textit{ZSON} \cite{majumdar2022zson}\\
            \specialrule{0.5pt}{0pt}{0pt}   
            \textbf{Repository} & \scriptsize{\url{https://github.com/gunagg/zson}} \\
            \textbf{Configs} & \scriptcode{configs/tasks/imagenav\_hm3d.yaml}\\
            \textbf{Link} &
            \href{https://github.com/gunagg/zson/blob/a0415137aeb36dab962b467bdfcfb51dbaa6ed71/configs/tasks/imagenav_hm3d.yaml#L13}{\scriptcode{github.com/gunagg/zson/blob/main/configs/tasks/imagenav\_hm3d.yaml\#L13}}\\
            \specialrule{0.5pt}{0pt}{0pt}
            \rowcolor{TableGray2}      
             \textbf{Model} & 
             \textit{PIRLNav} \cite{ramrakhya2023pirlnav}\\
            \specialrule{0.5pt}{0pt}{0pt}   
            \textbf{Repository} & \scriptsize{\url{https://github.com/Ram81/pirlnav}} \\
            \textbf{Configs} & \scriptcode{configs/experiments/il\_objectnav.yaml}\\
                    &$\hookrightarrow$  \scriptcode{configs/tasks/objectnav\_hm3d.yaml}\\
            \textbf{Link} &
            \href{https://github.com/Ram81/pirlnav/blob/8235b0e3b589818441f6783fecd5c4fa8ad53f1b/configs/tasks/objectnav_hm3d.yaml#L15}{\scriptcode{github.com/Ram81/pirlnav/blob/main/configs/tasks/objectnav\_hm3d.yaml\#L15}}\\
            \specialrule{0.5pt}{0pt}{0pt}
            \rowcolor{TableGray2}      
             \textbf{Model} & 
             \textit{GOAT-Bench} \cite{GoatBench2024}\\
            \specialrule{0.5pt}{0pt}{0pt}   
            \textbf{Repository} & \scriptsize{\url{https://github.com/Ram81/goat-bench}} \\
            \textbf{Configs} & \scriptcode{config/experiments/ver\_goat\_skill\_chain.yaml}\\
                    &$\hookrightarrow$  \scriptcode{config/tasks/goat\_stretch\_hm3d.yaml}\\
            \textbf{Link} &
            \href{https://github.com/Ram81/goat-bench/blob/74c41d19d4a4c3608d1575b512087b5a529aee0e/config/tasks/goat_stretch_hm3d.yaml#L78}{\scriptcode{github.com/Ram81/goat-bench/blob/main/config/tasks/goat\_stretch\_hm3d.yaml\#L78}}\\
            \specialrule{0.5pt}{0pt}{0pt}
            \rowcolor{TableGray2}      
             \textbf{Model} & 
             \textit{DEBiT} \cite{CrocoNav2024}\\
            \specialrule{0.5pt}{0pt}{0pt}   
            \textbf{Repository} & \scriptsize{\url{https://github.com/naver/debit}} \\
            \textbf{Configs} & \scriptcode{configs/imgnav-gibson-debit.yaml}\\
            \textbf{Link} &
            \href{https://github.com/naver/debit/blob/eecc8f9ba2d24c2e8a1ea9d1ef29bff629e4424f/configs/imgnav-gibson-debit.yaml#L15}{\scriptcode{github.com/naver/debit/blob/main/configs/imgnav-gibson-debit.yaml\#L15}}\\
            \specialrule{0.5pt}{0pt}{0pt}
            \rowcolor{TableGray2}      
             \textbf{Model} & 
             \textit{FGPrompt} \cite{sun2024fgprompt}\\
            \specialrule{0.5pt}{0pt}{0pt}   
            \textbf{Repository} & \scriptsize{\url{https://github.com/XinyuSun/FGPrompt}} \\
            \textbf{Configs} & \scriptcode{exp\_config/ddppo\_imagenav\_gibson.yaml}\\
                    &$\hookrightarrow$  \scriptcode{exp\_config/imagenav\_gibson.yaml}\\
            \textbf{Link} & \href{https://github.com/XinyuSun/FGPrompt/blob/4430fab50e74c817ab5c7bb28650e569566a234e/exp_config/base_task_config/imagenav_gibson.yaml}{N/A}\\
            \specialrule{1pt}{0pt}{0pt}
    \end{tabular}
\caption{\label{tab:sliding_repos} \textbf{How the \code{allow\_sliding} flag is set in other codebases.} Here we list, for each method, the repository address, the main configuration file used for the main experiment (with $\hookrightarrow$ we indicate the eventual config file composed with the main one where the \code{Sliding} flag is set), and the link to the line where the parameter is set. \textit{FGPrompt} does not appear to set \code{Sliding} to \code{False}, thus it uses the default value of \code{True}.}
}
\end{table*}
\section{Sliding setting in Habitat}

As discussed in the main paper, the \code{allow\_sliding} parameter, or \code{Sliding} in short, in the Habitat simulator impacts the way an agent behaves when colliding with obstacles: if set to \code{True} the agent slides along obstacles, while when \code{False} the agent stops. By default the flag is switched to \code{True}, but Kadian \etal~\cite{kadian20sm2real} showed that setting \code{Sliding=True} has a big detrimental impact in sim2real transfer. Therefore, it is common practice in the field to set \code{Sliding=False} to train and test agents in conditions that are as realistic as possible. 

We analyzed several publicly-available code repositories to assess how these set the \code{Sliding} flag. As already mentioned, by default it is \code{True}, and it is usually modified in a \code{yaml} configuration file. Habitat uses \textit{Hydra}\footnote{\url{https://hydra.cc/}} to manage the configuration, which allows to compose multiple configuration files, making it cumbersome sometimes to track the exact final configuration used for each experiment. In \cref{tab:sliding_repos} we list, for several relevant robot navigation methods using Habitat, the link to the repository, the configuration file used for the main experiment (and other config files composed with the main one indicated by $\hookrightarrow$), and a link to the line in the code where \code{allow\_sliding} is set to \code{False}. Furthermore, we have searched the repositories for occurrences of the string ``\code{allow\_sliding}'' to verify that it is not modified directly in the code. All methods but \textit{FGPrompt} set \code{Sliding} to \code{False}, hence this codebase uses the default setting of \code{allow\_sliding=True}. For completeness, we note that we could find an occurrence of \code{allow\_sliding=False} in a config file in the \textit{FGPrompt} repository,\footnote{\scriptsize{\url{https://github.com/XinyuSun/FGPrompt/blob/main/src/config/v2.yaml\#L11}}} but this file does not appear to be used anywhere in the code. As a final check, we installed and ran the code of \textit{FGPrompt} with the recommended settings and commands, and we could indeed verify that \code{allow\_sliding=True} in the Habitat environments used for training and testing.

\section{Additional experiments}

\myparagraph{Impact of GRU structure} For fairness of comparison, we used the same GRU architecture in all our experiments, with 2 layers and hidden state of size 128. In Table \ref{tab:debitgru} we show the results for training \textit{DEBiT-B} with the same GRU architecture reported in \cite{CrocoNav2024}, with one layer and hidden dimension 512. The experiment excludes any dependency on this choice: the results are very similar in spite of the different number of layers and the change in sizes of the recurrent memory $|\mathbf{h}_t|$.

\begin{table}[t] \centering
{\small
    \setlength{\tabcolsep}{2pt}
    \begin{tabular}{lcc|cc}
            \specialrule{1pt}{0pt}{0pt}
            \rowcolor{TableGray2}      
             \textbf{Model+backb.} & 
             \textbf{GRU architecture} &
             \textbf{Sliding} &
             \textbf{SR(\%)} & 
             \textbf{SPL(\%)}
             \\
            \specialrule{0.5pt}{0pt}{0pt}   
            \rowcolor{colorpretrain}
            \cellcolor{colorpretrainleg} DEBiT-B &
            \cellcolor{colorpretrainleg} 
            2 layers, $|\mathbf{h}_t|=128$ &
            \cellcolor{colorpretrainleg} 
            \no
            &
            81.7 & 52.0\\
            \rowcolor{colorpretrain}
            \cellcolor{colorpretrainleg} 
            DEBiT-B &
            \cellcolor{colorpretrainleg} 
            2 layers, $|\mathbf{h}_t|=128$ &
            \cellcolor{colorpretrainleg} 
            \yes
            &
            90.5 & 60.3\\
            \rowcolor{colorpretrain}
            \cellcolor{colorpretrainleg} DEBiT-B &
            \cellcolor{colorpretrainleg} 
            1 layer, $|\mathbf{h}_t|=512$ &
            \cellcolor{colorpretrainleg} 
            \no&
            83.7 & 54.8
            \\
            \rowcolor{colorpretrain}
            \cellcolor{colorpretrainleg} 
            DEBiT-B &
            \cellcolor{colorpretrainleg} 
            1 layer, $|\mathbf{h}_t|=512$ &
            \cellcolor{colorpretrainleg} 
            \yes&
            90.8 & 60.5
            \\
            \specialrule{1pt}{0pt}{0pt}
    \end{tabular}
\caption{\label{tab:debitgru} \textbf{Comparisons of \textit{DEBiT}} \cite{CrocoNav2024} in two settings: our standard 2-layer GRU test architecture used in the paper (Table 1), and the architecture originally reported in \cite{CrocoNav2024} (1-layer GRU).}
}
\end{table}

\end{document}